\documentclass[letterpaper, 10 pt, conference]{ieeeconf} 
\usepackage[utf8]{inputenc}
\usepackage{todonotes}
\usepackage{amsmath}
\usepackage{caption}
\usepackage{subcaption}
\usepackage{float}
\usepackage{multirow}
\usepackage{colortbl}
\usepackage{booktabs}
\usepackage{tabularx}
\usepackage{xcolor}

\long\def\invis#1{}  

\makeatletter
\def\bstctlcite{\@ifnextchar[{\@bstctlcite}{\@bstctlcite[@auxout]}}
\def\@bstctlcite[#1]#2{\@bsphack
  \@for\@citeb:=#2\do{%
    \edef\@citeb{\expandafter\@firstofone\@citeb}%
    \if@filesw\immediate\write\csname #1\endcsname{\string\citation{\@citeb}}\fi}%
  \@esphack}
\makeatother 

\title{CARTIER: Cartographic lAnguage Reasoning Targeted at Instruction Execution for  Robots
}
\author{Dmitriy Rivkin, Nikhil Kakodkar, Francois Hogan, Bobak H. Baghi, Gregory Dudek}

\definecolor{lightgray}{gray}{0.95}
\long\def\greg#1{{\bf \color{red} #1 --gd}}  
\long\def\fhogan#1{{\bf \color{blue} #1}}
\long\def\drivkin#1{{\bf \color{purple} #1}}
\long\def\nikhil#1{{\bf \color{orange} #1}}
\long\def\bhb#1{{\bf \color{teal} #1}}

\long\def\greg#1{}  
\long\def\fhogan#1{}
\long\def\drivkin#1{}
\long\def\nikhil#1{}
\long\def\bhb#1{}

\long\def\invis#1{}  
\newcommand\gderror[1]{
  \typeout{--------------------------------------------------------------------}
  \typeout{------- #1 ---------}
  \typeout{--------------------------------------------------------------------}
  {\bf #1}
}
\newcounter{gdTmp}
\newcounter{gdLastCount}
\newcommand\maxpage[2][Error]{  
\ifnum\value{page}>#2
    \gderror{\Large \bf On page {\thepage} we are past page #2 (too long).   #1 }
\else\fi
}
\newcommand\maxpageSinceLast[2][Error]{  
\ifnum \numexpr \value{page} - \value{gdLastCount}\relax>#2
    \gderror{Exceeds max length #2 pages. Page \thepage: #1}
\thepage\else\fi
\setcounter{gdLastCount}{\value{page}}
}

\begin{document}
\bstctlcite{BSTcontrol}
\maketitle

\begin{abstract}
This work explores the capacity of large language models (LLMs) to 
address problems at the intersection of spatial planning and natural language interfaces for navigation. We focus on following complex instructions that are more akin to natural conversation than traditional explicit procedural directives typically seen in robotics. Unlike most prior work where navigation directives are provided as simple imperative commands (e.g., ``go to the fridge"), we examine implicit directives obtained through conversational interactions.We leverage the 3D simulator AI2Thor to create household query scenarios at scale, and augment it by adding complex language queries for 40 object types. We demonstrate that a robot using our method CARTIER (Cartographic lAnguage Reasoning Targeted at Instruction Execution for Robots) can parse descriptive language queries up to 42\% more reliably than existing LLM-enabled methods by exploiting the ability of LLMs to interpret the user interaction in the context of the objects in the scenario.
\end{abstract}

\section{Introduction}
This paper explores the extent to which natural interaction is possible between human and robot in the context of a navigation task. We seek to answer the question: ``Can a robot infer its task in a navigational context without receiving an explicit command?'' 
Household robotic tasks are often  formulated using imperative commands with a template structure that can be abstracted as ``go-do'' commands (go somewhere, then do something), which are a narrow and
atypical subset of the traditional conversational interactions between humans~\cite{Amazonconvversational,karppi2019non}.
We explore to what extent a robotic assistant can execute 
commands based on conversational instructions akin to those used between humans?  Our existing voice assistants 
to date are so limited in this respect that the notion itself may seem astounding.

%
In this work, we focus primarily on unstructured variants of the first step (go somewhere) and measure the capacity of the robot to recognize the physical location associated with the intent of the query. While existing voice assistants and robotic
systems are capable of interpreting simple, straightforward specifications (e.g. ``get me a shirt"),  natural language between humans often uses complex descriptors that fall outside of the scope of current systems.
Endowing robots with the ability to  infer a goal from a natural specification is desirable as it provides a more effective mode of interaction for most users and it allows various navigation trade-offs to be optimized by the system (for example path length vs traction vs landmark visibility).


\begin{figure}
\includegraphics[width=0.48\textwidth]{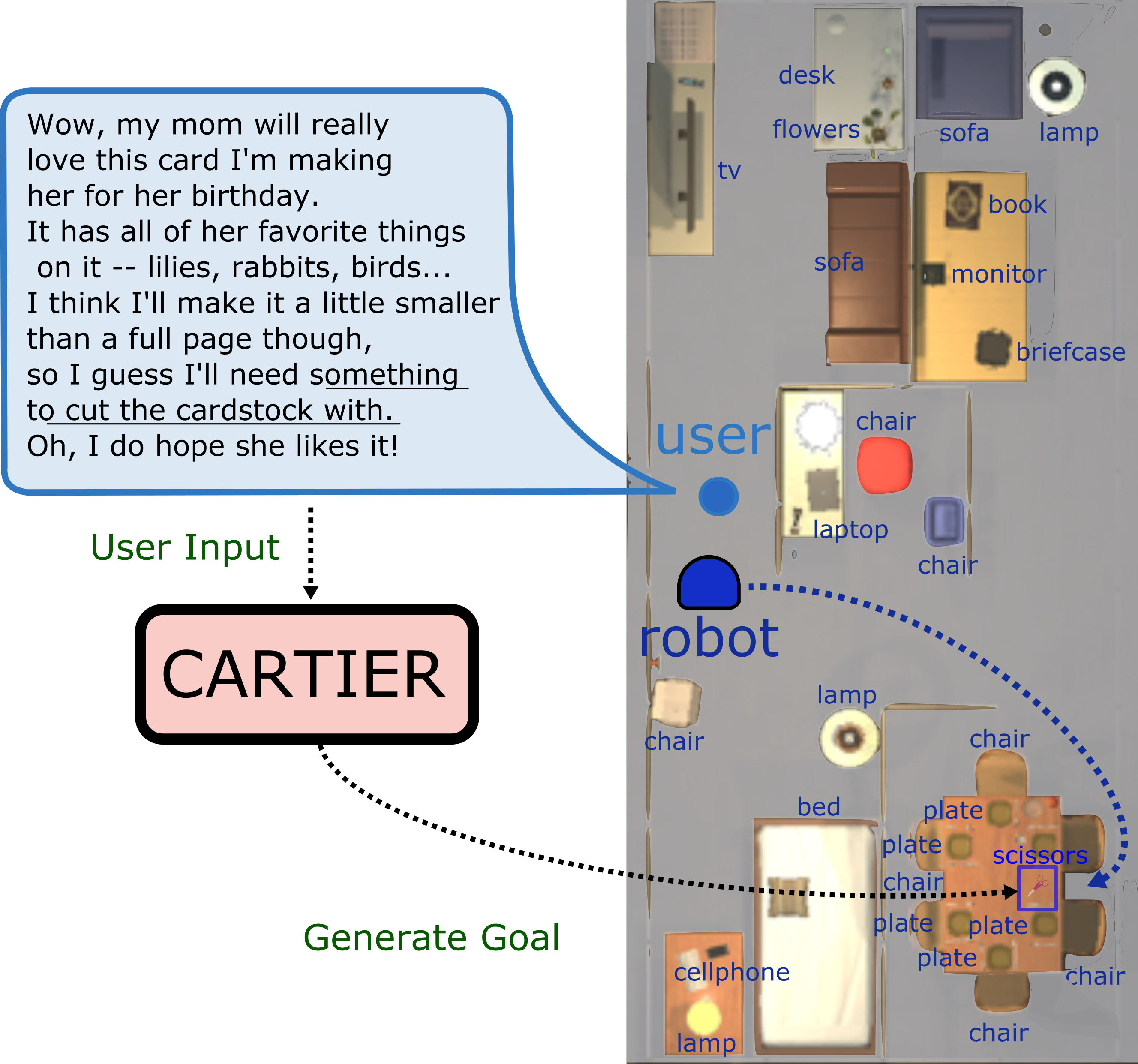}
\centering
\caption{CARTIER prompts an LLM with knowledge about a robot's environment in order to parse user intent from implicit, conversational queries. It then informs the robot where to navigate in order to help the user.}
\label{fig:query_types}
\end{figure}

For a robot, planning the navigation target requires a sophisticated understanding of the world which it inhabits. For example, consider the simple-seeming instruction ``Get a healthy snack." To effectively address this request, the robot must know what food the user has available (or at least where it can go to search for food), whether each food item qualifies as a snack, and how healthy each item is. Now imagine the user says something like: ``Am I hungry? I mean, am I really hungry or am I just bored? I don't know! I'm supposed to be on diet. And we just never have anything good to eat. How is it possible that we just got groceries delivered yesterday and we still have nothing good to eat? I think I'm hungry, and dinner is in like an hour."  As little as a year or two ago, the idea that a robot might be able to address such an interaction by navigating to an apple and bringing it to the user seemed unrealistic -- but by leveraging recent progress on several fronts, notably large language models (LLMs), this has become plausible. Significant challenges remain, however, in grounding the reasoning capabilities of LLMs in the reality of the physical world. The aim of this work is to examine how such challenges can be overcome in order to allow LLMs to be used to support such sophisticated interactions.

To study this question, we augment the AI2Thor \cite{ai2thor} simulation environment with implicit  conversational queries for $40$ object types. 
We then introduce CARTIER (Cartographic lAnguage Reasoning Targeted at Instruction Execution for Robots), a pipeline that first uses an LLM to infer which object the user is referring to in the query, then uses a ``spatial language index" to associate that object with a location in the scene. To tackle the problem of grounding the LLM in the robot's perception of its environment, we propose a novel method to caption the scene based on the objects the robot has detected within it by exploiting a pre-trained object detector. We then evaluate multiple approaches, of varying complexity, to construct and query the spatial language index.

The contributions of this paper are:
\begin{itemize}
    \item The augmentation of the AI2Thor simulator with three types of queries for $40$ object types: 1) Short explicit queries. 2) Short implicit queries where objects are referred to by their use rather than by their name in one sentence. 3) Long, implicit, and conversational queries that span several sentences, and include a significant amount of distracting content.
    \item The introduction of our method, CARTIER, that can handle all three query types. CARTIER consists of an offline stage where the robot explores the environment to build a spatial language index, and an online stage which processes user queries and relates them to physical points in space. The CARTIER reasoning pipeline is the first to address robot navigation by querying LLMs with holistic, object-centric representations.
    \item The evaluation of our method's performance in conjunction with several spatial language indices.
    \item A proof-of-concept experiment demonstrating the feasibility of CARTIER deployment in a physical robotic system.
\end{itemize}

\section{Related Work}

\subsection{Foundation Models for Open-Vocabulary Language Navigation}
To exploit the association between images and language, we make use of the  visual language model  CLIP~\cite{clip}. CLIP is an acronym for Contrastive Language-Image Pre-Training and it refers to an approach for training large-scale models that can encode the relationship between text and images.  The approach uses a transformer architecture trained on a large dataset of paired images and text and can subsequently predict whether an image and associated caption  are related or not. 

Finding and localizing objects and image features, especially in response to queries, is a  fundamental robotics problem which is often handicapped by a lack of prior knowledge~\cite{pateras1995understanding,sim2001learning,shridhar2018interactive,shridhar2020ingress}. Since the release of CLIP,  several authors have explored how to make use of this powerful vision-language model to enable language-driven navigation instructions using open vocabularies (i.e. not constrained to a fixed set of objects). Several works, including  VLMaps~\cite{vlmaps}, ConceptFusion~\cite{conceptfusion}, and Clip Fields~\cite{clipfields} generate per-pixel or per-voxel embeddings in the CLIP embedding space, which can then be queried using the CLIP language encoder. 
As a consequence of the related embedding spaces, however, these models have the same limitations as CLIP in terms of the sophistication of the query that they can successfully interpret. Some work, including VLMaps~\cite{vlmaps}, CLIP-nav~\cite{clipnav}, and LMNav~\cite{lmnav}, uses an LLM such as GPT-3~\cite{gpt3} in an effort to parse longer navigation instructions into a sequence of moves which can be executed by the robot. While these authors also make  use of an LLM to parse instructions, the nature of the instructions differs from  those considered in this paper as they   focus on a detailed description of the robot's path  rather than interpreting complex humans queries describing the desired location of the robot.
Clip on Wheels (CoW)~\cite{cows} presents a general framework for CLIP-based object navigation, where a robot navigates using some policy while attempting to localize the object in question using CLIP. As such, some of the papers mentioned earlier in this section can also be considered CoWs, although the precise problem formulation taken by the different authors differs. CARTIER is fundamentally different from these results due to its use of an LLM to interpret the user query. However, most of the work mentioned in this section could be used to implement the navigation component of CARTIER, once the correct object has been identified.

NLMap \cite{chen2023open} augments Say-Can \cite{brohan2023can} with mapping and navigation capabilities, using an open-vocabulary CLIP-based map similar to VLMaps. An LLM is used to make object proposals based on the user query, then the map is searched for the presence of these proposed objects. If any of the objects are located in the map, they are used in the construction of a planning prompt. FindThis~\cite{findthis} focuses on interactive dialogues which support localizing the specific instances of an object (e.g. ``the cup by the sink"). Similarly to NLMap, objects must first be proposed based on the query before they can be found in the environment. CARTIER differs from both in that it uses an object detector to compile a list of all objects in the scene, thereby providing more information to the planner.

Finally, ANSEL Photobot~\cite{ansel} used an LLM to break down enigmatic user queries into ensembles of concepts that can be detected using CLIP, however the focus was on open-vocabulary photo selection rather than navigation. 




\subsection{Captioning for Question Answering}
A key component of the CARTIER pipeline is to leverage an LLM to inform the robot on the contents of the scene. We do this by enumerating the objects present in the scene, having first detected them with a pre-trained object detector. Other work has also explored captioning images \cite{captionvqa} and $3$D scenes \cite{scan2cap}, \cite{sqa3d} in order to support visual question answering (VQA) and situated question answering (SQA), respectively. With all of these approaches, the way the scene is captioned constrains the types of questions that can be answered. SQA3D \cite{sqa3d} proposes to solve a situated question answering task using scene captioning and an LLM, but the captioning style is different from that used in CARTIER, necessitated by the difference in the target task. The task of interest in SQA3D has to do with answering queries related to the spatial arrangement of objects (e.g. ``what is behind the couch?").

There is good evidence that simple queries often yield unsatisfactory results and these can be enhanced via a more natural or conversational paradigm~\cite{keyvan2022approach,lopez2018alexa}.
A more complete coverage of the long history and relationship between natural and expressive query languages, and data collecting agents, most of which relates to Internet search is outside the scope of what we can cover here~\cite{lieberman2001exploring,hu2016natural,allen1983recognizing,lieberman1995letizia}.

\section{Dataset}
\label{sec:dataset}
The AI2Thor simulation environment contains $189$ different object types. We select a subset of $40$ objects and create three different types of queries (see Section \ref{sec:three_query_types} for details). Since the objects are distributed across many different scenes in different combinations, we can evaluate the same query across many different scenes. In this paper, we consider $7$ scenes over which we evaluate our approach. Not all objects are found in all scenes, so in total the dataset contains 127 (query, scene) pairs for each query type. For each (query, scene) pair, the dataset contains a list of objects contained in the scene which could plausibly satisfy the query. Using ground-truth object bounding boxes from the simulator, we compute the minimum distance between the predicted position and any of the bounding boxes of the objects. Using the distance as a performance metric rather than the object identity is convenient because it supports comparison to methods which do not explicitly output the identity of the object being searched for.
The environment exploration phase is achieved by teleoperating a mobile robot, which collects RGBD frames from a camera mounted on the robot at approximately $1.5$ meters from the ground. 

\subsection{Query Types}
\label{sec:three_query_types}
The queries are split into $3$ categories with increasing complexity: explicit, implicit, and conversational.  Explicit queries are straightforward requests to go to some explicitly-named object (e.g. ``get the scissors''), and can be successfully answered using an object detector or CLIP-based methods in the open vocabulary case. Implicit queries refer to an object implicitly by expressing some need of the user which could be addressed with that object, e.g. ``I need something to cut this paper with." The Implicit queries in our dataset are short (one or two short sentences) and to the point, focused entirely on discussing the need which is to be serviced. These cannot be answered with an object detector, since there is no straightforward way to extract the object of interest from the utterance, but in some cases may still be answered satisfactorily by CLIP, which is likely to learn to embed ``scissors" nearer to the phrase ``cut paper" than any other object. Our third class of queries, Conversational, are comprised of several sentences and may contain a significant amount of information that is irrelevant to the need expressed by the user. For example, consider the utterance depicted in Fig.~\ref{fig:query_types}. There are several challenges associated with using CLIP to answer this query directly. First, the length of CLIP's sentence embeddings is limited to $77$ tokens, meaning it may not even be possible to encode the entire text of the query. Second, CLIP will attempt to incorporate all of the concepts expressed in the utterance into the embedding (mom, birthday, lilies, rabbits, birds, etc), the majority of which are irrelevant to the need of the user. In order to interpret the query, and estimate  what the user is asking for, an LLM capable of processing lengthy pieces of text is required. These conversational queries are quite different from the manner in which people must speak to today's voice assistants. Based on the evidence from information retrieval, natural queries should enhance usability and effectiveness~\cite{keyvan2022approach}.



\begin{figure*}
\vspace{5mm}
\includegraphics[width=0.90\textwidth]{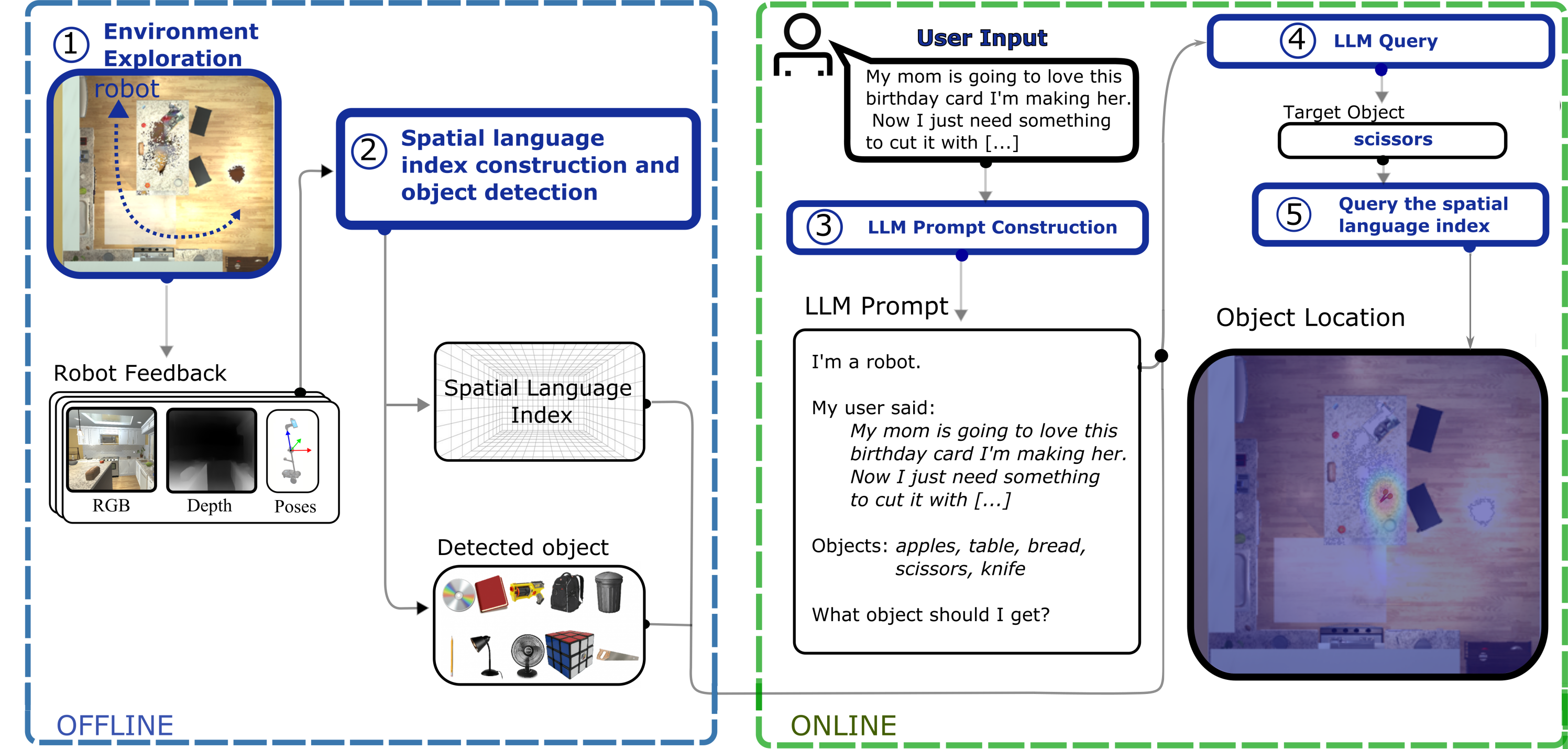}
\centering
\caption{Overview of CARTIER. Blue boxes indicate processes, while black boxes indicate data. \textbf{1.} The robot explores its environment during a preprocessing stage, and a trajectory of poses and RGBD frames is collected. \textbf{2.} In the next stage of preprocessing, an object detector is used to determine which objects are present in the scene. In addition, a spatial language index is built in order to support querying for locations given an object. We explore two such indices: VLMaps \cite{vlmaps} and a simpler solution which uses the bounding boxes returned by the object detector. \textbf{3.} User input text is combined with a list of object in the scene to produce an engineered prompt for the LLM \textbf{4.} The LLM is queried with the constructed prompt. We use ChatGPT and GPT-4 \cite{instructgpt} in our experiments. \textbf{5.} The spatial language index is used to look up the location of the object.}
\label{fig:pipeline}
\end{figure*}

\section{Method}
In this Section, we introduce CARTIER, a framework that can interpret complex user language instructions to generate navigation targets. CARTIER is composed of five parts: 1) Environment exploration, 2) Object detection and spatial language index construction, 3) LLM query construction, 4) Querying the LLM,  5) Querying the spatial language index. These are summarized graphically in Figure \ref{fig:pipeline}. The rest of the this section explores each of these parts in more detail.
\subsection{Environment exploration}
The robot navigates the environment collecting RGBD video, as well as the associated pose in which each video frame was collected. In the simulated environment, the robot pose is obtained from the simulator directly. 
Depth information and 6-DOF pose is required in order to maximize the accuracy of object localization, however certain versions of CARTIER, presented in the following sections, can work using only a monocular camera and 2-DOF pose estimates.

\subsection{Object detection and spatial language index construction}
\label{sec:object_detection_and_spatial_index}
CARTIER populates the LLM query with a list of objects in the scene, so we use an off-the-shelf object detector (EVA \cite{EVA}) trained on the LVIS dataset \cite{lvis} that contains $1204$ object categories. We run the detector on each of the frames in the video and compile a list of all LVIS objects encountered by the robot (with confidence score $> 0.8$) during the exploration phase. Though we elect to use an object detector with a fixed vocabulary, CARTIER could potentially handle open vocabulary queries. See Section \ref{sec:query_the_llm} for more discussion of this point.

A ``spatial language index" is a map that allows for a lookup of positions given an object name. We explore three such indices in this work. In order of descending order of complexity, the first is VLMaps \cite{vlmaps} which creates a 2-D top-down grid of the environment, where each grid cell represents a CLIP embedding (calculated via LSeg \cite{lseg}) averaged over each point projected into the grid. The second involves selecting the frame with the largest bounding box for the given object (a heuristic for proximity to the object), then using the depth signal in combination with the object bounding box to determine the position of the object. More specifically, the average position (i.e. ``center of mass'') for depth measurements within the bounding box is used as an estimate of object location.  The third method returns the position of the robot at the time that the frame with the largest object bounding box was observed. The benefit of this method is that it does not require a depth camera. We improve the accuracy of our proximity heuristic (maximize bounding box size) by partially compensating for perspective distortion using a transform based on the weak perspective camera model. 


We refer to these methods as VLMaps, ObjectDepth, and ObjectViewpoint, respectively.

\subsection{LLM query construction}
\label{sec:query_construction}
We wish for the LLM to select the most relevant object in the scene that fits the user query. The LLM is informed of the contents of the scene via the inclusion of a comma separated list of the objects that were detected. Each object type is only listed once, even if multiple instances were detected. In order to condition the correct behavior, the LLM is prompted with a short description of the task as follows:

\begin{figure}[H]
\includegraphics[width=0.46\textwidth]{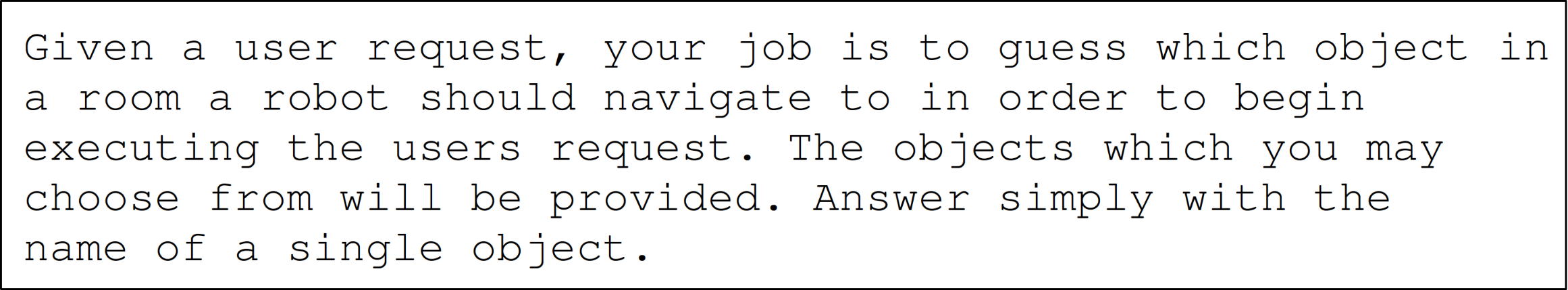}
\centering
\end{figure}


This prompt is targeted at LLMs fine-tuned for instruction following using RLHF (Reinforcement Learning from Human Feedback), such as ChatGPT and GPT-4. Unlike prompts for earlier text completion models (such as GPT-3) which often require several examples of the desired behavior in order to achieve performance \cite{ansel}, this prompt is terse, containing only a description of the desired behavior. By keeping the prompt
short and high-level, we avoid the risk of over-engineering the model's behavior to a particular case, leaving space for the model to apply its own best judgement to answer the query.
\drivkin{Frank and Greg: does this sound good?}

\subsection{Query the LLM}
\label{sec:query_the_llm}
We query the LLM with temperature set to 0 in order to sample maximum likelihood tokens and minimize the stochasticity of our results. In our experiments, we use ChatGPT and GPT-4. We do not specify stop tokens. Empirically we found that both models were reliable in terms of answering each query with a single object. In some cases, some minimal parsing was required to extract the identity of the objects due to responses such as ``In order to fulfil the user request, the robot should navigate to the object called `Bed'".

Because CARTIER forces the LLM to select from a fixed set of objects known from the object detector, one could erroneously assume that CARTIER is a closed-vocabulary method. For an illustrative example of why this is not true, consider a query in which the user wants a tamper (an object used to compress coffee when making espresso), but ``tamper" is not contained in the vocabulary of the object detector. An LLM can use its significant world-knowledge to infer that the tamper is likely to be found near the coffee machine, and direct the robot to the correct location. In fact, even if the tamper had been hidden from the view of the robot during the exploration phase, it would still likely find the right answer, as long as a reasonable guess can be made about where to find the object in question.

\subsection{Query the spatial language index}
In order to query the VLMaps index, we embed the object name using CLIP and then select the pixel with the highest similarity, after first dropping the large embedding dimensions following \cite{ansel}. The ObjectDepth and ObjectViewpoint indices are queried as described in Section \ref{sec:object_detection_and_spatial_index}.

\section{Evaluation}
We evaluate CARTIER by running it on the dataset described in Sec \ref{sec:dataset} using both ChatGPT and GPT-4 as the LLM. We measure both the correctness of the LLM output and the distance to the target object achieved by the spatial language index. We compare against two recently published baselines. Details are provided below.

\subsection{Baselines}
We consider two baselines, which we refer to VLMaps-baseline and NLMap. VLMaps-baseline consists of querying the VLMaps spatial language index with the user query directly. If the user query is longer than 77 tokens, it is truncated. The NLMap baseline uses the prompt from \cite{chen2023open} in order to generate object proposals. The generated proposals are then queried against the VLMaps spatial language index, and the similarity value to the nearest-match grid square for each proposal is thresholded (threshold value of 12.05) to generate the list of object proposals found in the scene. These proposals passed to the LLM along with the CARTIER prompt described in Section \ref{sec:query_construction} in order to select the final object. This method can only be used with the VLMaps spatial language index, as it produces open vocabulary outputs which are not guaranteed to align with the vocabulary of the object detector used to implement ObjectDepth and ObjectViewpoint.

\subsection{Metrics}
\label{sec:metrics}
Each query in the dataset has an associated list of objects in the scene which may be reasonable answers to the query. The first metric (referred to as object-match) measures the correctness of the text response produced by the LLM, i.e. whether the object it has responded with is in the list. This evaluation is performed manually, as the automation of evaluation presents two challenges. The first is the presence of synonyms (e.g. ``trash can", ``trash bin", ``garbage can", etc.) in natural language. This is not such an issue for CARTIER, which selects object from a fixed vocabulary, but presents more of a challenge for evaluating NLMap, an open-vocabulary method. The second is co-location of multiple objects. For example, the target object may be a ``sink'' but the LLM selects ``faucet.'' This should be counted a success, since the mapping between sinks and faucets is (usually) 1:1. On the other hand, if the target object is ``bed" but the LLM selects ``sheets" this is a failure, since sheets are sometimes co-located with beds, but are also sometimes found in closets. Manual evaluation allows for a case-by-case approach to these matters. Note that metric is not applied to the VLMaps baseline as it does not produce a text-based answer.

The second metric (referred to Euclidian-distance) measures the distance (in meters) between the inferred object location and the nearest ground-truth object in the scene. This metric avoids any issues related to object co-location -- if the sheets are indeed on the bed, both answers are equally good. The downside is that it conflates the performance of the LLM and the spatial language index, therefore the two metrics are complementary.

\section{Results}

\begin{table*}[htbp]
\vspace{3mm}
    \centering
    \renewcommand{\arraystretch}{1.5}
    \begin{tabular}{|>{\centering\arraybackslash}m{1.8 cm}|>{\centering\arraybackslash}m{1.8cm}|>{\centering\arraybackslash}m{1.8cm}|>{\centering\arraybackslash}m{2.2cm}|>{\centering\arraybackslash}m{1.8cm}|>{\centering\arraybackslash}m{1.8cm}|>{\centering\arraybackslash}m{1.8cm}|}
        \hline
        \rowcolor{lightgray}
        \textbf{Model} & \textbf{Type} & \textbf{CARTIER}  & \textbf{CARTIER} & \textbf{CARTIER} & \textbf{NLMap} & \textbf{VLMaps}\\
        \rowcolor{lightgray}
                \textbf{} & \textbf{} & \textbf{ObjectDepth} & \textbf{ObjectViewpoint} & \textbf{VLMaps}& \textbf{} & \textbf{Baseline}  \\
        \hline
gpt4 & conversational& \textbf{1.20} & 1.94 & 1.70  & 2.11 & 2.35 \\ \hline
chatgpt & conversational & \textbf{1.39} & 2.04 & 1.78 &  2.28 & 2.35 \\ \hline
gpt4 & implicit & \textbf{1.17} & 1.89 & 1.66  & 2.09 & 2.27 \\ \hline
chatgpt & implicit & \textbf{1.20}& 1.80 & 1.69  & 2.22 & 2.27 \\ \hline
gpt4 & explicit & \textbf{1.09} & 1.91 & 1.70  & 2.02 & 1.93 \\ \hline
chatgpt & explicit & \textbf{1.21}& 1.96 & 1.71  & 1.89 & 1.93 \\ \hline
    \end{tabular}
\caption{\label{tab:mean_results} 
Mean (over all scenes and tasks) distances, in meters, between predicted object location and ground truth object location for a collection of different methods, question types, and language models.
}
\end{table*}

\begin{figure}
\includegraphics[width=0.38\textwidth]{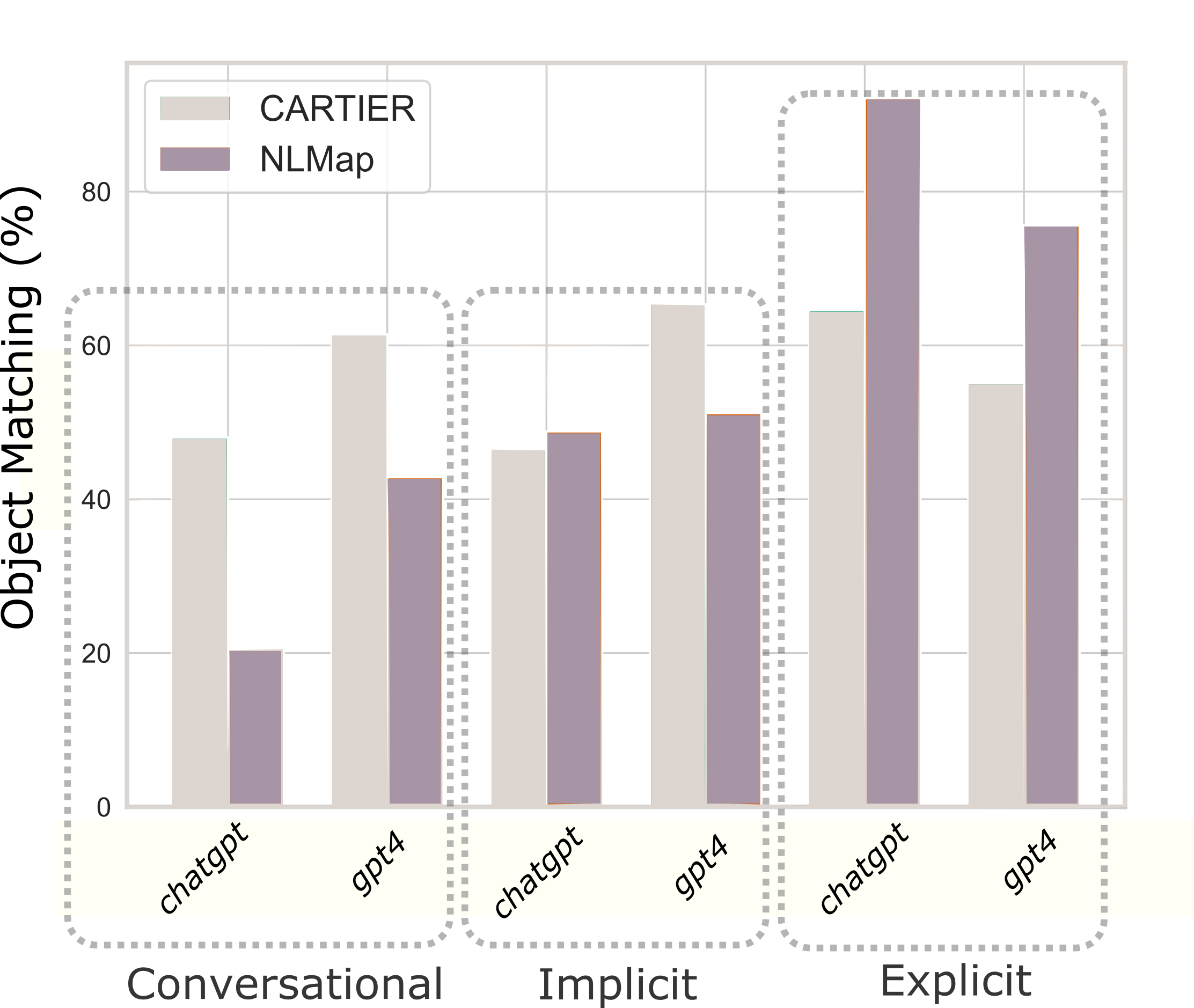}
\centering
\caption{Comparison of object-matching performance of CARTIER and NLMap across different models and query types. The reported score is the average success rate over all queries and environments.}
\label{fig:object_matching}
\end{figure}

The results according to the Euclidian-distance metric are presented in Table \ref{tab:mean_results}, while Figure \ref{fig:object_matching} summarizes results for the object-match metric. 
We observe that CARTIER is significantly stronger than other baselines according to the euclidean-distance metric, and also that this gap is more pronounced for the conversational queries, which are the most challenging. The ObjectDepth spatial language index appears to be significantly better at localizing objects than VLMaps, suggesting that open-vocabulary mapping is not always necessary. ObjectViewpoint was the weakest spatial language index, but this is not surprising because it is only expected to be accurate when the robot has passed very near the object on its exploration trajectory.

Figure \ref{fig:object_matching} also confirms the superiority of CARTIER on the challenging conversational queries. CARTIER's performance using GPT4 on all three query types is comparable, suggesting that the limiting factor is the object detector, rather than the reasoning capabilities of the LLM. NLMap beats CARTIER on explicit queries in terms of the object-matching metric, but loses on euclidean-distance. This suggests that CARTIER is able to exploit object co-location (as described in Sections \ref{sec:query_the_llm}  \ref{sec:metrics}) and/or more robustly localize the objects it does find. For an illustration of this effect, consider ``bed-sheets'' example from Section ~\ref{sec:metrics}. In many scenes, bed and sheets are likely to indeed be co-located, so while the manual object-matching score may consider the answer ``sheets" to be wrong answer, the distance-based metric is likely to make no distinction between the two.

On the whole, these results help to confirm the underlying CARTIER hypothesis -- as queries become more complex and nuanced, contextualizing the LLM-based decision making process withe details about the contents of the scene increases the LLM's ability to correctly resolve the query.

\section{Real-world deployment}

\begin{figure}\includegraphics[width=0.45\textwidth]{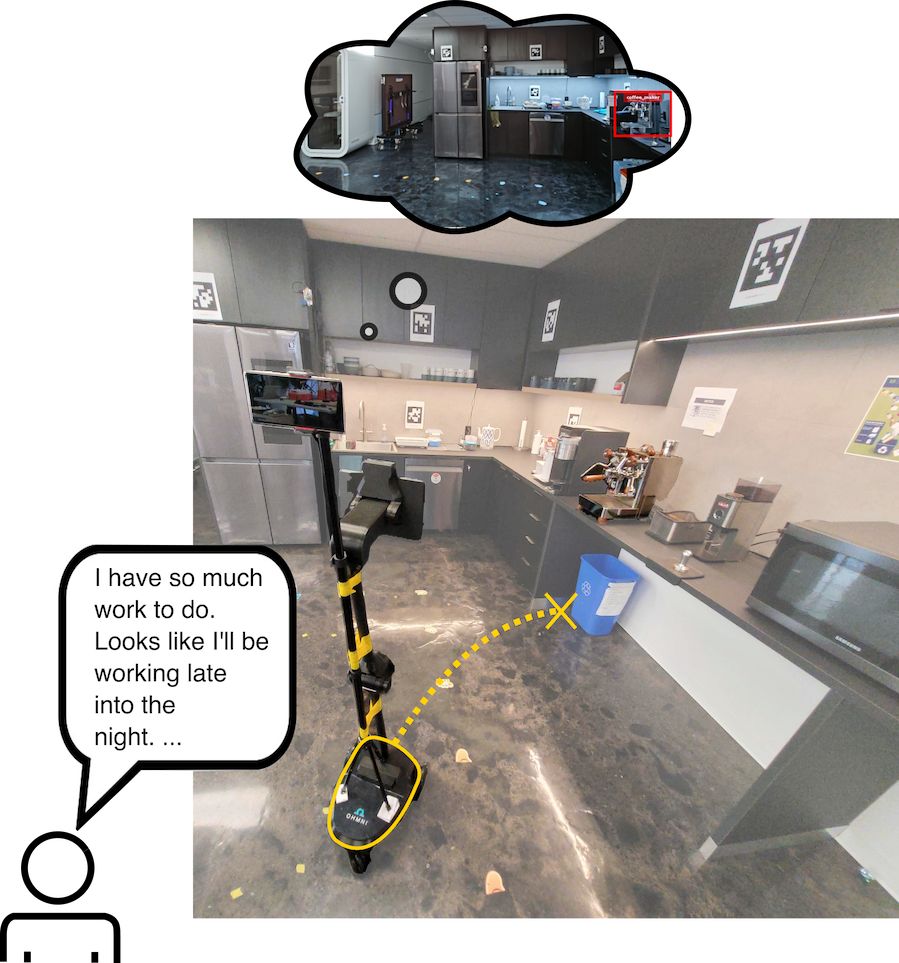}
\centering
\caption{Real-world deployment of CARTIER. Following a conversational query, the robot can navigate to the location (coffee machine) that best satisfies the user's intent.}
\label{fig:real_world}
\end{figure}

In this section, we deploy  CARTIER in a real-world scenario where a telepresence robot (Ohmni by Ohmnilabs) is given human queries and must navigate to the appropriate location. In Fig.~\ref{fig:real_world}, a robot is given the conversational query ``I have so much work to do. Looks like I'll be working late into the night. I am guessing the presentation would take me another two hours or so to finish. But I am getting really sleepy. We better do something about that.'' Using CARTIER, the robot determines that within all the objects detected in the environment, the coffee machine best matches the human's intent and can navigate to the coffee machine's position using CARTIER's perception pipeline. In this proof-of-concept, we generate the exploration trajectory by teleoperating the robot during which we record $394$ images using a Samsung S23 Ultra mobile phone. The pose trajectory of the robot and the depth images are generated from the sequences of RGB frames using the software package ARCore by Google.

\section{Conclusion}
The existing work on the value and impact of conversation search for the Internet~\cite{keyvan2022approach} suggests that 
users are also likely to desire more natural, conversational interactions with robot assistants in the near future. In this work, we took a step towards enabling those types of interactions by introducing CARTIER, a method which fuses power of modern LLMs and state of the art object detectors. 

CARTIER's approach of providing the LLM decision maker with a detailed description of the scene contents represents an alternative approach to those described in work such as NLMap, where the LLM must first hypothesise which objects are useful given the input query. Our results show that this approach provides more relative benefit as the complexity of the queries increases. One potential downside is that the object detector may ignore certain object types, or fail to distinguish important object attributes, such as whether a plate is dirty. Thus, an interesting avenue for future work may be to synthesise the two approaches by both listing known objects in the prompt and allowing the LLM to ask a VLM targeted questions about the objects. By allowing users to specify ``go-do tasks'' in more natural ways, we hope that the utility and impact of robot navigation agents will be greatly enhanced.

\maxpage[Six page limit before bibliography. ``The page limit is 6 pages for the paper (text, figures, tables, acknowledgement, etc.) + any number of pages for the bibliography/references. Papers exceeding the (6+n) page limit at the time of submission will be returned without review.'']{6}

\bibliographystyle{IEEEtran}
\bibliography{IEEEabrv,biblio}

\begin{thebibliography}{10}
\providecommand{\url}[1]{#1}
\csname url@rmstyle\endcsname
\providecommand{\newblock}{\relax}
\providecommand{\bibinfo}[2]{#2}
\providecommand\BIBentrySTDinterwordspacing{\spaceskip=0pt\relax}
\providecommand\BIBentryALTinterwordstretchfactor{4}
\providecommand\BIBentryALTinterwordspacing{\spaceskip=\fontdimen2\font plus
\BIBentryALTinterwordstretchfactor\fontdimen3\font minus
  \fontdimen4\font\relax}
\providecommand\BIBforeignlanguage[2]{{%
\expandafter\ifx\csname l@#1\endcsname\relax
\typeout{** WARNING: IEEEtran.bst: No hyphenation pattern has been}%
\typeout{** loaded for the language `#1'. Using the pattern for}%
\typeout{** the default language instead.}%
\else
\language=\csname l@#1\endcsname
\fi
#2}}

\bibitem{Amazonconvversational}
``What is conversational ai?''
  \url{https://developer.amazon.com/alexa-skills-kit/conversational-ai},
  accessed: 2023-09-14.

\bibitem{karppi2019non}
T.~Karppi and Y.~Granata, ``Non-artificial non-intelligence: Amazon’s alexa
  and the frictions of ai,'' \emph{Ai \& Society}, vol.~34, pp. 867--876, 2019.

\bibitem{ai2thor}
E.~Kolve, R.~Mottaghi, W.~Han, E.~VanderBilt, L.~Weihs, A.~Herrasti, D.~Gordon,
  Y.~Zhu, A.~Gupta, and A.~Farhadi, ``{AI2-THOR: An Interactive 3D Environment
  for Visual AI},'' \emph{arXiv}, 2017.

\bibitem{clip}
\BIBentryALTinterwordspacing
A.~Radford, J.~W. Kim, C.~Hallacy, A.~Ramesh, G.~Goh, S.~Agarwal, G.~Sastry,
  A.~Askell, P.~Mishkin, J.~Clark, G.~Krueger, and I.~Sutskever, ``Learning
  transferable visual models from natural language supervision,'' 2021.
  [Online]. Available: \url{https://arxiv.org/abs/2103.00020}
\BIBentrySTDinterwordspacing

\bibitem{pateras1995understanding}
C.~Pateras, G.~Dudek, and R.~De~Mori, ``Understanding referring expressions in
  a person-machine spoken dialogue,'' in \emph{1995 International Conference on
  Acoustics, Speech, and Signal Processing}, vol.~1.\hskip 1em plus 0.5em minus
  0.4em\relax IEEE, 1995, pp. 197--200.

\bibitem{sim2001learning}
R.~Sim and G.~Dudek, ``Learning environmental features for pose estimation,''
  \emph{Image and Vision Computing}, vol.~19, no.~11, pp. 733--739, 2001.

\bibitem{shridhar2018interactive}
M.~Shridhar and D.~Hsu, ``Interactive visual grounding of referring expressions
  for human-robot interaction,'' \emph{arXiv preprint arXiv:1806.03831}, 2018.

\bibitem{shridhar2020ingress}
M.~Shridhar, D.~Mittal, and D.~Hsu, ``Ingress: Interactive visual grounding of
  referring expressions,'' \emph{The International Journal of Robotics
  Research}, vol.~39, no. 2-3, pp. 217--232, 2020.

\bibitem{vlmaps}
C.~Huang, O.~Mees, A.~Zeng, and W.~Burgard, ``Visual language maps for robot
  navigation,'' \emph{arXiv preprint arXiv:2210.05714}, 2022.

\bibitem{conceptfusion}
\BIBentryALTinterwordspacing
K.~M. Jatavallabhula, A.~Kuwajerwala, Q.~Gu, M.~Omama, T.~Chen, S.~Li, G.~Iyer,
  S.~Saryazdi, N.~Keetha, A.~Tewari, J.~B. Tenenbaum, C.~M. de~Melo,
  M.~Krishna, L.~Paull, F.~Shkurti, and A.~Torralba, ``Conceptfusion: Open-set
  multimodal 3d mapping,'' 2023. [Online]. Available:
  \url{https://arxiv.org/abs/2302.07241}
\BIBentrySTDinterwordspacing

\bibitem{clipfields}
\BIBentryALTinterwordspacing
N.~M.~M. Shafiullah, C.~Paxton, L.~Pinto, S.~Chintala, and A.~Szlam,
  ``Clip-fields: Weakly supervised semantic fields for robotic memory,'' 2022.
  [Online]. Available: \url{https://arxiv.org/abs/2210.05663}
\BIBentrySTDinterwordspacing

\bibitem{clipnav}
V.~S. Dorbala, G.~Sigurdsson, R.~Piramuthu, J.~Thomason, and G.~S. Sukhatme,
  ``Clip-nav: Using clip for zero-shot vision-and-language navigation,''
  \emph{arXiv preprint arXiv:2211.16649}, 2022.

\bibitem{lmnav}
D.~Shah, B.~Osinski, B.~Ichter, and S.~Levine, ``Lm-nav: Robotic navigation
  with large pre-trained models of language, vision, and action,'' \emph{arXiv
  preprint arXiv:2207.04429}, 2022.

\bibitem{gpt3}
T.~Brown, B.~Mann, N.~Ryder, M.~Subbiah, J.~D. Kaplan, P.~Dhariwal,
  A.~Neelakantan, P.~Shyam, G.~Sastry, A.~Askell, S.~Agarwal, A.~Herbert-Voss,
  G.~Krueger, T.~Henighan, R.~Child, A.~Ramesh, D.~Ziegler, J.~Wu, C.~Winter,
  C.~Hesse, M.~Chen, E.~Sigler, M.~Litwin, S.~Gray, B.~Chess, J.~Clark,
  C.~Berner, S.~McCandlish, A.~Radford, I.~Sutskever, and D.~Amodei, ``Language
  models are few-shot learners,'' in \emph{Advances in Neural Information
  Processing Systems}, H.~Larochelle, M.~Ranzato, R.~Hadsell, M.~Balcan, and
  H.~Lin, Eds., vol.~33.\hskip 1em plus 0.5em minus 0.4em\relax Curran
  Associates, Inc., 2020, pp. 1877--1901.

\bibitem{cows}
\BIBentryALTinterwordspacing
S.~Y. Gadre, M.~Wortsman, G.~Ilharco, L.~Schmidt, and S.~Song, ``Cows on
  pasture: Baselines and benchmarks for language-driven zero-shot object
  navigation,'' 2022. [Online]. Available:
  \url{https://arxiv.org/abs/2203.10421}
\BIBentrySTDinterwordspacing

\bibitem{chen2023open}
B.~Chen, F.~Xia, B.~Ichter, K.~Rao, K.~Gopalakrishnan, M.~S. Ryoo, A.~Stone,
  and D.~Kappler, ``Open-vocabulary queryable scene representations for real
  world planning,'' in \emph{2023 IEEE International Conference on Robotics and
  Automation (ICRA)}.\hskip 1em plus 0.5em minus 0.4em\relax IEEE, 2023, pp.
  11\,509--11\,522.

\bibitem{brohan2023can}
A.~Brohan, Y.~Chebotar, C.~Finn, K.~Hausman, A.~Herzog, D.~Ho, J.~Ibarz,
  A.~Irpan, E.~Jang, R.~Julian, \emph{et~al.}, ``Do as i can, not as i say:
  Grounding language in robotic affordances,'' in \emph{Conference on Robot
  Learning}.\hskip 1em plus 0.5em minus 0.4em\relax PMLR, 2023, pp. 287--318.

\bibitem{findthis}
\BIBentryALTinterwordspacing
A.~Majumdar, F.~Xia, brian ichter, D.~Batra, and L.~Guibas, ``Findthis:
  Language-driven object disambiguation in indoor environments,'' in \emph{7th
  Annual Conference on Robot Learning}, 2023. [Online]. Available:
  \url{https://openreview.net/forum?id=nNsZxc2cmO}
\BIBentrySTDinterwordspacing

\bibitem{ansel}
\BIBentryALTinterwordspacing
D.~Rivkin, G.~Dudek, N.~Kakodkar, D.~Meger, O.~Limoyo, X.~Liu, and F.~Hogan,
  ``Ansel photobot: A robot event photographer with semantic intelligence,''
  2023. [Online]. Available: \url{https://arxiv.org/abs/2302.07931}
\BIBentrySTDinterwordspacing

\bibitem{captionvqa}
\BIBentryALTinterwordspacing
S.~Changpinyo, D.~Kukliansky, I.~Szpektor, X.~Chen, N.~Ding, and R.~Soricut,
  ``All you may need for vqa are image captions,'' 2022. [Online]. Available:
  \url{https://arxiv.org/abs/2205.01883}
\BIBentrySTDinterwordspacing

\bibitem{scan2cap}
D.~Z. Chen, A.~Gholami, M.~Nießner, and A.~X. Chang, ``Scan2cap: Context-aware
  dense captioning in rgb-d scans,'' 2020.

\bibitem{sqa3d}
\BIBentryALTinterwordspacing
X.~Ma, S.~Yong, Z.~Zheng, Q.~Li, Y.~Liang, S.-C. Zhu, and S.~Huang, ``Sqa3d:
  Situated question answering in 3d scenes,'' 2022. [Online]. Available:
  \url{https://arxiv.org/abs/2210.07474}
\BIBentrySTDinterwordspacing

\bibitem{keyvan2022approach}
K.~Keyvan and J.~X. Huang, ``How to approach ambiguous queries in
  conversational search: A survey of techniques, approaches, tools, and
  challenges,'' \emph{ACM Computing Surveys}, vol.~55, no.~6, pp. 1--40, 2022.

\bibitem{lopez2018alexa}
G.~L{\'o}pez, L.~Quesada, and L.~A. Guerrero, ``Alexa vs. siri vs. cortana vs.
  google assistant: a comparison of speech-based natural user interfaces,'' in
  \emph{Advances in Human Factors and Systems Interaction: Proceedings of the
  AHFE 2017 International Conference on Human Factors and Systems Interaction,
  July 17- 21, 2017, The Westin Bonaventure Hotel, Los Angeles, California, USA
  8}.\hskip 1em plus 0.5em minus 0.4em\relax Springer, 2018, pp. 241--250.

\bibitem{lieberman2001exploring}
H.~Lieberman, C.~Fry, and L.~Weitzman, ``Exploring the web with reconnaissance
  agents,'' \emph{Communications of the ACM}, vol.~44, no.~8, pp. 69--75, 2001.

\bibitem{hu2016natural}
R.~Hu, H.~Xu, M.~Rohrbach, J.~Feng, K.~Saenko, and T.~Darrell, ``Natural
  language object retrieval,'' in \emph{Proceedings of the IEEE conference on
  computer vision and pattern recognition}, 2016, pp. 4555--4564.

\bibitem{allen1983recognizing}
J.~Allen, ``Recognizing intention from natural language utterances,''
  \emph{Computational Model of Discourse}, pp. 107--166, 1983.

\bibitem{lieberman1995letizia}
H.~Lieberman \emph{et~al.}, ``Letizia: An agent that assists web browsing,''
  \emph{IJCAI (1)}, vol. 1995, pp. 924--929, 1995.

\bibitem{instructgpt}
\BIBentryALTinterwordspacing
S.~Changpinyo, D.~Kukliansky, I.~Szpektor, X.~Chen, N.~Ding, and R.~Soricut,
  ``All you may need for vqa are image captions,'' 2022. [Online]. Available:
  \url{https://arxiv.org/abs/2205.01883}
\BIBentrySTDinterwordspacing

\bibitem{EVA}
Y.~Fang, W.~Wang, B.~Xie, Q.~Sun, L.~Wu, X.~Wang, T.~Huang, X.~Wang, and
  Y.~Cao, ``Eva: Exploring the limits of masked visual representation learning
  at scale,'' \emph{arXiv preprint arXiv:2211.07636}, 2022.

\bibitem{lvis}
A.~Gupta, P.~Dollar, and R.~Girshick, ``{LVIS}: A dataset for large vocabulary
  instance segmentation,'' in \emph{Proceedings of the {IEEE} Conference on
  Computer Vision and Pattern Recognition}, 2019.

\bibitem{lseg}
B.~Li, K.~Q. Weinberger, S.~Belongie, V.~Koltun, and R.~Ranftl,
  ``Language-driven semantic segmentation,'' \emph{arXiv preprint
  arXiv:2201.03546}, 2022.

\end{thebibliography}

\end{document}